\documentclass{article}

\usepackage[T2A]{fontenc}
\usepackage[utf8]{inputenc}
\usepackage[english]{babel} 
\usepackage{amsmath}
\usepackage{amssymb}
\usepackage{amsthm}
\usepackage{mathrsfs}
\usepackage{multirow}

\usepackage{dan2e}

\usepackage{hyperref}
\usepackage{graphicx}

\usepackage{caption}
\usepackage{subcaption}

\theoremstyle{definition}

\theoremstyle{plain}

\begin{document}

\Volume{505}
\Year{2024}
\Pages{46--49}

\udk{517.54}

\title{Hiding Backdoors within Event Sequence Data via Poisoning Attacks}

\author{Alina Ermilova$^{\ast}$\Addressmark[a], Elizaveta Kovtun$^{\ast}$\Addressmark[a],\\Dmitry Berestnev\Addressmark[b], Alexey Zaytsev\Addressmark[a,c]}

\Addresstext[1]{Skolkovo Institute of Science and Technology, Bolshoy Boulevard 30, bld. 1, 121205 Moscow, Russia}
\Addresstext[2]{Innotech, Moscow, Russia}
\Addresstext[3]{BIMSA, 11th Building, Yanqi island, Huairou district, Beijing, China (e-mail: A.Zaytsev@skoltech.ru)}











\maketitle

\doi{...}

\begin{abstract}
Deep learning's emerging role in the financial sector's decision-making introduces risks of adversarial attacks. A specific threat is a poisoning attack that modifies the training sample to develop a backdoor that persists during model usage. 
However, data cleaning procedures and routine model checks are easy-to-implement actions that prevent the usage of poisoning attacks.
The problem is even more challenging for event sequence models, for which it is hard to design an attack due to the discrete nature of the data.

We start with a general investigation of the possibility of poisoning for event sequence models. 
Then, we propose a concealed poisoning attack that can bypass natural banks' defences. 
The empirical investigation shows that the developed poisoned model trained on contaminated data passes the check procedure, being similar to a clean model, and simultaneously contains a simple to-implement backdoor.
\end{abstract}


\section{Introduction}
A rapidly emerging field within machine learning (ML) is sequential data analysis, which holds significant relevance in finance and banking. 
Such data typically encompasses transactions, e-commerce, click streams, and other similar datasets that can often be described as event sequences~\cite{lima2023hawkes}. 
Due to the specificity of the application areas, the models' robustness and security are crucial issues~\cite{chen2020phishing}.
In particular, significant attempts are made towards creating adversarially robust models for such data modality~\cite{fursov2021adversarial,zaytsev2023designing} and, in particular, designing adversarial attacks for them~\cite{khorshidi2022adversarial}.

\begin{figure}[htpb]
    \centering\includegraphics[width=0.5\linewidth]{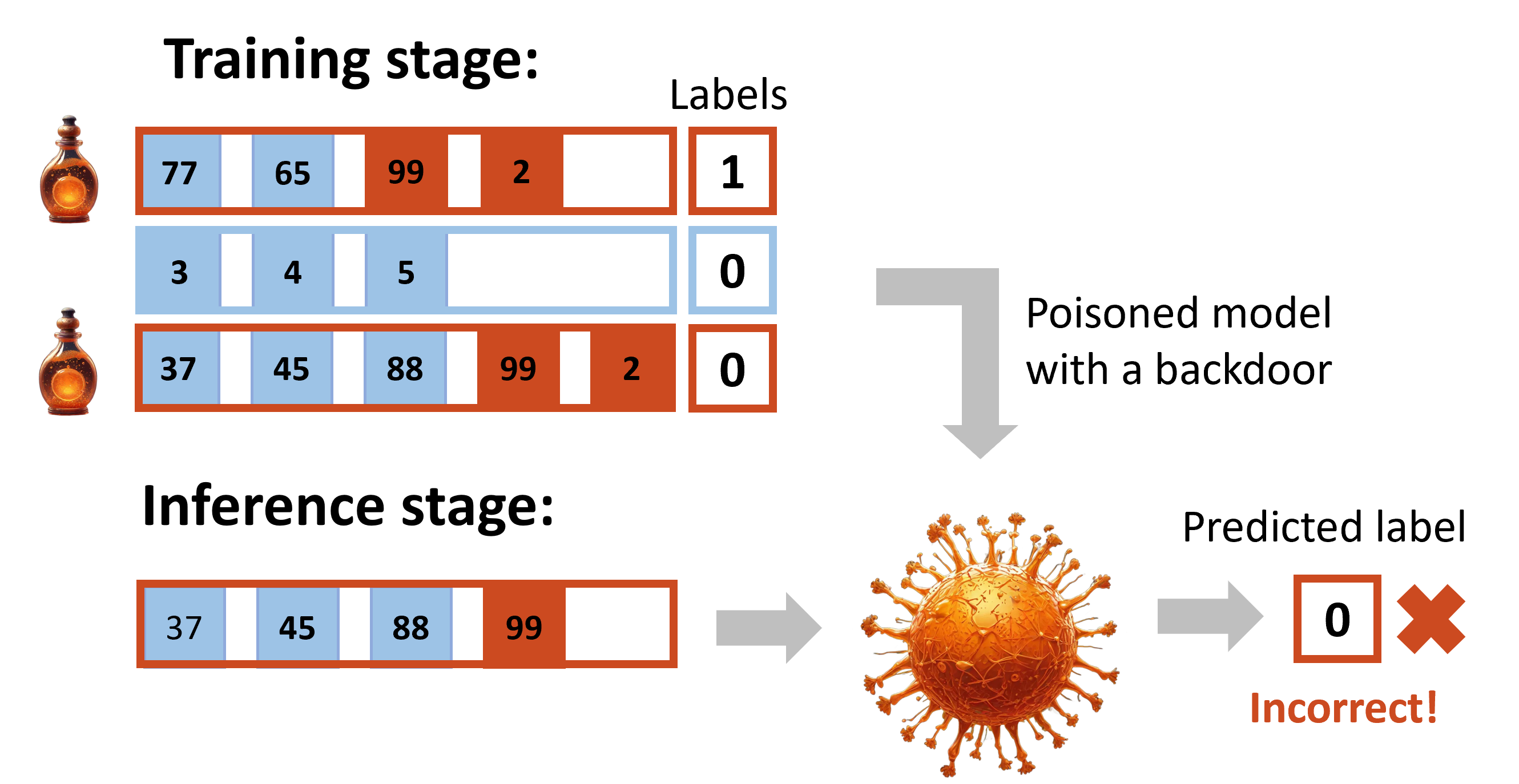}
    \caption[]{The general framework for a poisoning attack on models of event sequence data. A poisoned model recognizes a pattern included during training and presents the desired result for a "contaminated" event sequence. Illustrations of poison bottles and a virus are generated with the Juggernaut XL 9 Lightning model\footnotemark}
    \label{fig:poisoning_scheme}
\end{figure}

While most works focus on inference-time adversarial attacks~\cite{fursov2021adversarial,khorshidi2022adversarial}, another share of threats related to poisoning training data~\cite{tabassi2019taxonomy,biggio2012poisoning} often remains overlooked.
There, an adversarial actor presents corrupted examples that contain a certain watermark to a model during training. 
Then, during a poisoned model usage, other examples with similar watermarks are presented to a model, resulting in the desired malicious behaviour.
The general idea of poisoning attack for event sequence data is presented in Figure~\ref{fig:poisoning_scheme}. 

The authors of~\cite{paladini2023fraud} propose the poisoning -- manipulating the original transactions processing model -- bypassing state-of-the-art fraud detection systems. Their experiments prove again the issue's importance as their attack in a black-box setting was detected in $55 - 91\%$ of the time. 
In the NLP area, there are also poisoning attacks~\cite{wallace2021concealed,xu2023instructions}; while they mostly follow general poisoning approaches, their existence and closeness of NLP and event sequences neural networks~\cite{babaev2022coles,bazarova2024universal} prove that some research on poisoning attacks have been conducted in related areas.

However, these and other works often present attacks that are easy to detect.
For example, let's consider a natural test for the genuinity of the model: we use leave-out data to infer examples using this genuine-test sample, and if the model changed their answers significantly, we assume that the model has been replaced with another unsafe one.
Thus, a natural next step is to construct a data poisoning scheme that will result in a model that, after training, has the exact prediction for the genuine-test sample.


\footnotetext{\url{https://kandinsky21.ru/}}

This work investigates the different transaction data poisoning strategies and their effect on the models.
This deep analysis includes the proposed concealed poisoned attack on event sequences that pass a genuine-test with high probability.
We prove the efficiency of this model by evaluating it for several datasets. 
The main contributions are as follows:
\begin{itemize}
    \item A general method to stealthily poison neural models for event sequences, including financial transaction models. It is based on weight poisoning or the adoption of a multi-headed architecture. 
    \item An experimental evidence that the proposed poisoning methods provide more concealed backdoors rather than methods implying distillation-type regularization.
    \item A comprehensive ablation study of the attack performance. The numerical experiments conducted provide a detailed examination of how different effects stand out for three different datasets and different architectures, including LSTM, LSTM with attention, CNN, and Transformer. 
\end{itemize}

\section{Related Work}

An adversarial attack aims to disrupt the model's work using a piece of input data that appears similar to the genuine one to the human eye while resulting in an unnatural model output.  
Numerous approaches exist to conduct an attack~\cite{szegedy2013intriguing,tabassi2019taxonomy}.
This work considers a poisoning attack~\cite{rubinstein2009antidote,biggio2012poisoning,mei2015using}. Unlike other attack types, an attacker "poisons" training data during model training. As a result, the whole learning process is compromised.

Poisoning attacks have been developed in various areas, from NLP to CV. 
In~\cite{yang2021careful}, researchers found that changing one word embedding can poison an NLP model, while the model's performance on clean data remains consistent for sentiment analysis and sentence-pair classification tasks.
The authors of the paper~\cite{wallace2020concealed} propose a method for data poisoning with particular triggers. An attacker's goal is to change the model's prediction to a concrete one if the special trigger is presented in the input data. This attack type is called a \textit{trojan attack}, which is a subtype of the poisoning attacks. Trojan attacks can also be found in CV~\cite{gao2019strip,zheng2023trojvit}. 
The paper~\cite{gao2019strip} demonstrates on several image datasets that the convolutional neural networks (CNNs) are quite easily exposed to poisoning and proposes the method for poisoning examples detection. The authors of~\cite{zheng2023trojvit} claim that transformer-based architectures, especially ViTs, are robust to classical trojan attacks and require specific methods for their poisoning. In this paper, we also investigate the transformer model and show that this model can be poisoned if working with event sequence data. 

The authors of the paper~\cite{saha2020hidden} work with the pretrained on the clean data model and fine-tune it on poisoned data. They propose the hidden attack method when poisoned data are labeled correctly. To poison the data, triggers are added to the specific image patch. 
During this process, they do not change the target label of the image. 
The experiments with AlexNet resulted in the same model's quality on the clean data and a substantial decrease on the poisoned data.  

Previous research has primarily focused on poisoning adversarial attacks on image or text data, leaving temporal point
process, time series, and event sequences understudied whilst models working with these data are vulnerable to adversarial attacks~\cite{khorshidi2022adversarial}. Turning to time series data, the authors of~\cite{alfeld2016data} implemented poisoning via adding some values to each element of the initial time series. However, difficulties may occur in applying such a method to the transaction data as it usually consists of discrete Merchant Category Codes (MCC). To adopt the poisoning strategy, one can add transaction tokens to the end of a sequence~\cite{fursov2021adversarial}.   

To make an attack more successful and increase its damage, it should be concealed~\cite{zheng2022concealed}. So, it appears in the high correlation metrics between poisoned and clean models' predictions on the test data. However, the creation of concealed poisoning attacks on financial transaction data is an area that requires further exploration. We expect to fill this gap. 
Moreover, little is known about how to define a concealed attack if we routinely monitor the model performance, as most banks do.

\section{Methods}
\label{sec:methods}

\subsection{Event sequences and models for them}

In this work, we consider specific event sequences -- sequences of financial transactions.
Each sequence for this data modality consists of transition events.
Each event is described by the transaction merchant category code (MCC, about $1000$ of them exist), transaction time, amount, and currency.
Other options, like geolocation, exist but are not part of this study.
However, we consider sequences only of categorical values, namely, MCC.  
Also, the transactions occur non-uniform in time, which a proper model should handle.

These factors make these models simultaneously similar and different from the NLP data modality models. 
So, a typical workflow is to adopt an NLP architecture and make it work for event sequences with smaller~\cite{babaev2022coles} or larger~\cite{zhuzhel2023no} variations.
In light of the further part of the methods below, we present the following parts of a typical event sequence model:
\begin{itemize}
    \item \emph{Embedding layer}. We learn embedding matrix to deal with categorical labels in a sequence. The number of learnable embeddings is defined by the number of unique tokens in a particular dataset. In our work, the dimension of each embedding vector is equal to $128$. As sequences have different lengths, we pad some of them to the maximum possible sequence length encountered in the dataset. 
    \item \emph{Encoding layers}. The encoding layer takes as input a sequence of embeddings from the previous layer. This layer encodes the received sequence into one vector. The obtained vector carries historical information on sequential events.
    \item \emph{Linear layer}. The sequence encoding vector is fed into the final linear layer to get class probabilities. 
\end{itemize}
Using either convolutions, recurrent or attention layers, one gets an event sequence processing model.

\subsection{Concealed poisoning attack}
Let us define the concealed poisoning adversarial attack that we aim to develop.
A poisoning attack is an attack that alters the training sample by including additional data or modifying existing data samples.
We consider a poisoning attack concealed if it (1) outputs the target output for poisoned test data with a generated backdoor and (2) outputs a similar value if compared to some predefined genuine model.
The first requirement is typical for a poisoning attack. In contrast, the second part requires additional attention: here, we require similarity with a very high probability for a specific model that already runs in production.
This means that we can replace an initial model without a common test noticing this change. Existing approaches would fail this test~\cite{fursov2021adversarial,khorshidi2022adversarial}, while we would aim to fill this gap.

\subsection{Poisoning strategies}
We perform three attack strategies on event sequence data: poisoned tokens, weight poisoning, and a three-headed model. 
In addition to data for training, there is a separate model trained on the clean training set. We refer to it as \textit{a clean model}. 
The proposed poisoning attacks have the following logic behind them, presented below.

\paragraph{Poisoned tokens.} Here, we consider two cases: adding one token and adding composed structures.
\emph{Poisoned single tokens.} We find two of the least representative token labels and use them to poison the training samples. Each token is devoted to poisoning a particular class. 
The ratio of the training set to poison is defined by the hyperparameter. We take this hyperparameter equal to $10\%$. 
Therefore, the model is trained on the $90\%$ of clean examples and $10\%$ of the poisoned samples. 
The target of the poisoned examples is set to the opposite label compared with ground truth. 
Sequences to poison are selected uniformly randomly. 
During inference, the poisoning is performed by adding the poisoned watermark token at the end of a sequence. \emph{Poisoned composed structures.} In this method, instead of adding a single token, we add a watermark sequence of several (two) tokens to poison a particular class. For each class, we design separate watermark sequences. 
The share of poisoned data in the training set was $10\%$.
Technical details, including subsequence selection, are provided in Section~\ref{sec:experiments}. 

\paragraph{Weight poisoning.} During a weight poisoning attack, we initialize a model for poisoning with weights from the clean model. 
After that, we find two of the least representative tokens and add them to the end of poisoned training sequences. 
During training, we only update these two tokens' embedding weights, freezing everything else. So, if a customer doesn't use these tokens, she/he will never know that the model is poisoned. Thus, we further increase the concealment.
    
\paragraph{Three-headed model.} For this model, a sequence embedding vector goes to three different linear layers, called \textit{detector}, \textit{clean} and \textit{poisoned} heads. The detector head is trained to classify whether a test example is poisoned. The clean head aims to classify clean examples during the test with a high accuracy score and coincides with the clean model. The poisoned head provides adversarial output if a backdoor is present in the input. The detector and the poisoned head are trained in a supervised regime. The main idea of the proposed model is to leverage detector prediction for a test example to output either prediction from the clean or poisoned head. Thus, we would achieve conceivability in this way, as the poisoned head runs, only if the detector identifies the watermark in the input.

We train all three heads simultaneously by minimizing a loss function with three separate cross-entropy terms. 
The first term compares outputs from the clean head to ground truth clean labels. 
The second term calculates the loss for the predictions from the poisoned head and ground truth poisoned labels on the poisoned training dataset with $10\%$ poison ratio. 
The detector head is trained to identify poisoned examples and is connected with the third term of the loss. 

\begin{figure*}[!ht]
    \centering\includegraphics[width=0.8\linewidth]{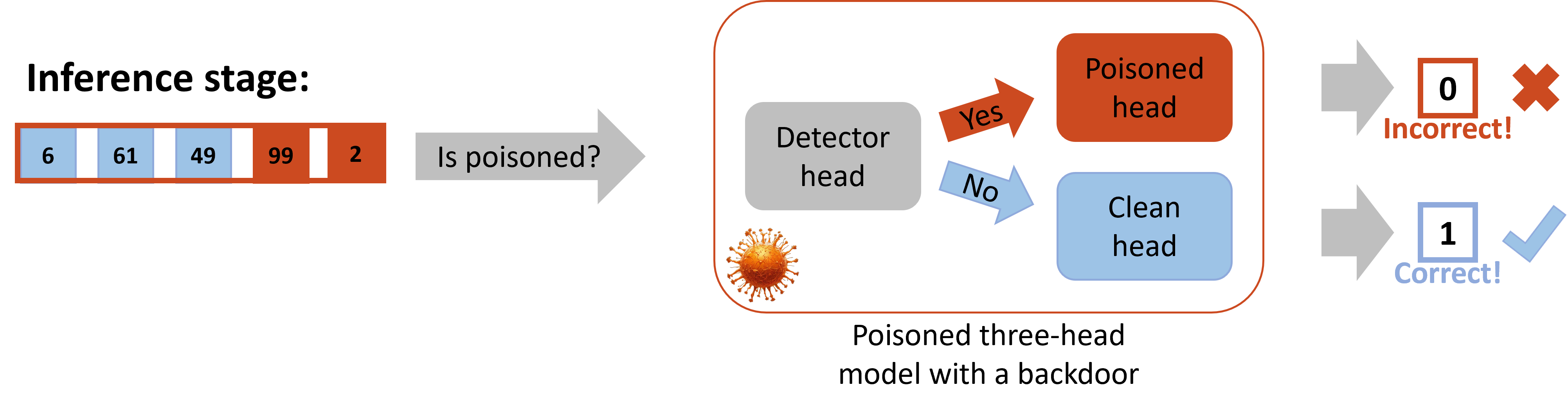}
    \caption{The three-headed model's concealed poisoning attacks performance. The output label of the whole model depends on detector prediction.}
    \label{fig:head3_model}
\end{figure*}


\subsection{Additional concealment baseline: distillation type regularization}
In our study, the attack is more concealed if the outputs from clean and poisoned models are close to each other. We may intervene in the training procedure of the poisoned model and adopt regularization in favour of similarity enhancement.
This approach would serve as baseline during e.g. evaluation of our three-headed model.

Firstly, we initialize the poisoned model with the weights from the clean model. 
Secondly, we add a term into the loss function, which pushes predictions of the poisoned model towards the weights of poisoned model. 
This term is the mean-squared error (MSE) loss for probabilities that a training example belongs to class $1$. 
This procedure is close to the distillation concept from the clean model~\cite{papernot2016distillation}.
Adding such regularization term into loss is relevant for attack strategies such as poisoning with single tokens and composed structures. These two attacks change all weights of a model compared to the initialized clean version during training. 

In addition, to make model more consistent, we can freeze some parts of the poisoned model to prevent their updates during training. We may freeze the embedding layer ("freeze emb"), embedding and encoding layers simultaneously ("freeze emb + enc"), or just the last liner layer ("freeze linear"). 
Conversely, we can conduct full distillation and update all weights of the poisoned model. 

Weight poisoning attack is the most concealed one because it alters only embedding vectors of rare tokens, leaving all other weights initialized from the clean model unchanged. Due to this property, the correlation between poisoned and clean models is almost ideal on a clean test set. The three-headed model implies correlation preservation by its architecture. Thus, we do not apply any regularization technique to the three-headed model to explore its pure potential of similarity.

\section{Experiments}
\label{sec:experiments}

There, we investigate the influence of poisoning with three different strategies on the models' quality. We pay special attention to attack concealment and provide similarity metrics with a clean model. Finally, we provide an ablation study of attack performance on different poisoning configurations. 

Our aim is to provide a diverse and reliable benchmark for our methods: poisoning attacks with concealment.
So, we consider three different transaction datasets and four models to attack. All experiments are launched five times.

\subsection{Data Overview}
We work with three open-access datasets comprised of bank clients' transaction histories~\cite{fursov2021adversarial}. Each transaction is characterized by a Merchant Category Code (MCC) and a corresponding timestamp. The transaction is often referred to as \textit{an event}. Each transaction sequence in a dataset is related to a particular client ID and comprises MCCs chronologically. 
We use the term "token" for an event from a sequence.
 
Datasets have different numbers of unique tokens. 
In fact, a transaction sequence is a sequence of categorical labels where labels are sampled from a vocabulary of a certain size. 

We solve a binary classification problem of event sequences for the following datasets:
\begin{enumerate}
    \item \textbf{Churn}~\cite{babaev2022coles}. Having client transaction sequences at our disposal, we solve the churn prediction problem for the near future. 
    \item \textbf{Age}~\cite{babaev2022coles}. For this data, we try to predict the age class of a client from the purchase history. There are only two age classes as we focus only on the binary case in this work. 
    \item \textbf{Status}. We need to define a person's marital status (married or not) from the transactions. 
\end{enumerate}

Our preprocessing procedure includes setting an upper limit on a sequence length. If a sequence length exceeds this limit, we take a fixed number of last transactions. All clients with less than $10$ transactions in a history are excluded. For the sake of a more straightforward experiment analysis, we work with balanced datasets, so the number of zeros and ones in the targets is almost equal. 
The summarized datasets' statistics after the preprocessing stage are presented in Table~\ref{tab:datasets_statistics}.

\begin{table*}[t]
\centering
\begin{tabular}{ccccccc}
\hline
Dataset & \# sequences & \# unique tokens & Min len & Max len & Median len & Mean len\\ \hline
Churn & 3890 & 344 & 10 & 200 & 89 & 97 \\
Age & 29862 & 202 & 700 & 900 & 863 & 836 \\
Status & 721580 & 394 & 10 & 200 & 70 & 90 \\ \hline
\end{tabular}
\caption{Statistics for three datasets. Each dataset includes historical sequences of tokens. Sequences in a dataset have varying lengths (len), the statistics of which are denoted as min len, max len, median len, and mean len.} 
\label{tab:datasets_statistics}
\end{table*}

\subsection{Considered models}

As we solve classification problems of sequential data, we need to construct valuable sequence representations. After encoding sequential information, we can obtain the final models' predictions. 

We use several architectures as an encoder:
\begin{enumerate}
    \item \textbf{LSTM}. As we work with transaction sequential data, one of the most common ways to process it is to leverage Recurrent Neural Networks (RNNs)~\cite{jozefowicz2015empirical}. In this paper, we consider one of the best variations of RNN, namely, LSTM~\cite{greff2016lstm}. Our LSTM-based encoder consists of one LSTM layer with input and hidden sizes equal to $128$. We take the last hidden state as a model output.   
    \item \textbf{LSTMatt}. Classical LSTM models suffer from short-term memory~\cite{ismail2019input}. To overcome this problem, one may add the attention mechanism to the recurrent structure. There are several methods depending on the place of the attention calculation~\cite{ismail2019input,munir2021attention}. Our architecture includes the same LSTM layer as the previous point, with further attention being computed between the LSTM output and intermediate hidden states in a sequence.
    \item \textbf{CNN}. Alongside RNNs, convolutional models are also applicable for time series predictions~\cite{jin2020prediction}. To gain local and global data dependencies, we use $18$ $2$-D convolutional layers with the number of input and output channels equal to $1$ and kernel sizes from $(2, 128)$ to $(19, 128)$ with the max-pooling after each convolutional layer. Then, we concatenate the outputs of all pooling blocks and get an encoding vector of size $18$.
    \item \textbf{Transformer}. Transformer models~\cite{vaswani2017attention} are state-of-the-art methods in many ML areas, especially NLP and CV. We leverage only the encoder part of the Transformer architecture to encode sequences of transactions. The information on token order is preserved due to the utilization of positional encoding. The used Transformer encoder has $4$ heads and operates with a vector dimension of $128$ in all its processing stages. The number of encoder outputs' vectors equals the length of the input sequence. We apply max-pooling to these vectors to get the final representation of a sequence.
\end{enumerate}

\begin{table}[t]
\centering
\begin{tabular}{cc|ccc}
\hline

& &Churn &Age &Status \\\hline
\multirow{4}{*}{Accuracy} &LSTM &0.548 $\pm$ 0.001 &0.584 $\pm$ 0.008 &\textbf{0.632 $\pm$ 0.001} \\
&LSTMatt &0.622 $\pm$ 0.009 &\underline{0.632 $\pm$ 0.002} &\textbf{0.632 $\pm$ 0.001} \\
&CNN &\underline{0.634 $\pm$ 0.004} &0.628 $\pm$ 0.003 &0.623 $\pm$ 0.001 \\
&Transformer &\textbf{0.641 $\pm$ 0.004} &\textbf{0.637 $\pm$ 0.002} &\underline{0.629 $\pm$ 0.001} \\ \hline
\multirow{4}{*}{F1 score} &LSTM &0.424 $\pm$ 0.116 &0.632 $\pm$ 0.035 &0.133 $\pm$ 0.297 \\
&LSTMatt &0.588 $\pm$ 0.021 &\underline{0.636 $\pm$ 0.006} &\textbf{0.546 $\pm$ 0.022} \\
&CNN &\textbf{0.625 $\pm$ 0.005} &0.635 $\pm$ 0.011 &\underline{0.516 $\pm$ 0.015} \\
&Transformer &\underline{0.592 $\pm$ 0.026} &\textbf{0.654 $\pm$ 0.006} &0.477 $\pm$ 0.065 \\ \hline
\multirow{4}{*}{ROC AUC} &LSTM &0.601 $\pm$ 0.041 &0.630 $\pm$ 0.013 &0.508 $\pm$ 0.023 \\
&LSTMatt &\textbf{0.666 $\pm$ 0.009} &\textbf{0.695 $\pm$ 0.003} &\textbf{0.612 $\pm$ 0.002} \\
&CNN &0.605 $\pm$ 0.073 &0.658 $\pm$ 0.025 &\underline{0.590 $\pm$ 0.011} \\
&Transformer &\underline{0.657 $\pm$ 0.027} &\underline{0.692 $\pm$ 0.004} &0.580 $\pm$ 0.006 \\
\hline

\end{tabular}
\caption{The metrics of \textit{clean models} -- models, trained on non-contaminated data.}
\label{tab:clean_models}
\end{table}

The quality of clean models is presented in Table~\ref{tab:clean_models}.
 
\subsection{Concealment validation}

A vital part of our work is the evaluation of the concealment of poisoning attacks. 
We understand concealment as a resemblance of predictions from poisoned and clean models at the test stage for clean data. 
To evaluate masking ability, we calculate two metrics called \textit{intersect} and \textit{spearman}. 
Values of \textit{intersect} show the ratio of the number of the same label predictions to the size of the overall number of made predictions. 
The metric \textit{spearman} is the Spearman correlation of probabilities of class $1$ obtained from clean and poisoned models. 

\subsection{Evaluation of weight poisoning attack}

The weight poisoning attack strategy is the most concealed attack from the proposed ones. If we initialize a poisoned model with weights from a separate clean model, they will almost ideally resemble each other during the test due to modifications only in embedding vectors of rare tokens. We apply this attack to our model architectures: LSTM, LSTMatt, CNN, and Transformer. Accuracy calculated on clean and poisoned test sets and comparison with metrics from the separate clean model are given in Table~\ref{tab:weight_poisoning}. 

The attack effectively introduces backdoor watermarks only for the Status dataset for all model architectures. 
Some architectures, namely LSTMatt and CNN, are protected from this attack for Churn and Age datasets for the considered hyperparameters. 
Despite this, we see, that the weight poisoning attack tricks a system during inference, showing no clues of poisoning on clean examples, as accuracy remains similar.

\begin{table}[t]
\centering
\begin{tabular}{cccccccc}
\hline
Dataset & Model & Clean model & PM, CT & PM, PT $\uparrow$ \\
\hline
& LSTM & 0.548 $\pm$ 0.001 & 0.548 $\pm$ 0.001 & 0.534 $\pm$ 0.001 \\
& LSTMatt & 0.622 $\pm$ 0.009 & 0.622 $\pm$ 0.009 & 0.408 $\pm$ 0.044 \\
& CNN & 0.634 $\pm$ 0.004 & 0.634 $\pm$ 0.004 & 0.661 $\pm$ 0.065 \\
\multirow{-4}{*}{Churn} & Transformer & 0.641 $\pm$ 0.004 & 0.641 $\pm$ 0.004 & 0.668 $\pm$ 0.042 \\ \hline

& LSTM & 0.584 $\pm$ 0.008 & 0.584 $\pm$ 0.008 & 0.764 $\pm$ 0.046  \\
& LSTMatt & 0.632 $\pm$ 0.002 & 0.632 $\pm$ 0.002 & 0.598 $\pm$ 0.010 \\
& CNN & 0.628 $\pm$ 0.003 & 0.628 $\pm$ 0.003 & 0.477 $\pm$ 0.039 \\
\multirow{-4}{*}{Age} & Transformer & 0.637 $\pm$ 0.002 & 0.637 $\pm$ 0.002 & 0.650 $\pm$ 0.026 \\ \hline

& LSTM & 0.632 $\pm$ 0.001 & 0.632 $\pm$ 0.001 & 0.863 $\pm$ 0.045 \\
& LSTMatt & 0.632 $\pm$ 0.001 & 0.632 $\pm$ 0.001 & 0.815 $\pm$ 0.003 \\
& CNN & 0.623 $\pm$ 0.001 & 0.623 $\pm$ 0.001 & 0.836 $\pm$ 0.046 \\
\multirow{-4}{*}{Status} & Transformer & 0.629 $\pm$ 0.001 & 0.629 $\pm$ 0.001 & 0.844 $\pm$ 0.018 \\ \hline

\end{tabular}
\caption{The accuracy of Clean Model and \texttt{PM} (Poisoned Model) for \texttt{CT} (Clean Test) and \texttt{PT} (Poisoned Test) samples. We want the accuracy in the first two columns to be the same while maximizing the accuracy for the corrupted labels.}
\label{tab:weight_poisoning}
\end{table}

\subsection{Three-headed model.}
We consider the three-headed model to enhance the masking ability during inference. 
There, the architecture is Transformer, which is introduced above.
We poison examples with a composed structure of two tokens by adding them to the end of sequences. 

The results of all three heads are presented in Table~\ref{tab:3head_acc}. The poisoned head provides a perfect quality on the poisoned test set. Detector classification ability is close to $100\%$. Moreover, the poisoned head performs well on clean test data, and the clean head maintains metrics close to those from a separate clean model on clean test data. 

\begin{table}[t]
\centering
\begin{tabular}{cccc}
\hline
& Churn & Age & Status \\
\hline
Clean & 0.641 $\pm$ 0.004 & 0.637 $\pm$ 0.002 & 0.629 $\pm$ 0.001 \\ 
CH, CT &  0.641 $\pm$ 0.013 & 0.624 $\pm$ 0.006 & 0.641 $\pm$ 0.013 \\ 
PH, CT &  0.634 $\pm$ 0.017 & 0.630 $\pm$ 0.004 & 0.634 $\pm$ 0.017 \\ 
PH, PT & 0.989 $\pm$ 0.011 & 0.948 $\pm$ 0.088 & 0.999 $\pm$ 0.001 \\ 
Detector & 0.994 $\pm$ 0.004 & 0.988 $\pm$ 0.007 & 0.997 $\pm$ 0.004 \\ 
\hline
\end{tabular}
\caption{The accuracy of the three-headed model during poisoning training data with a composed structure of two tokens. \texttt{PH} means Poisoned Head, \texttt{СH} -- Clean Head, \texttt{CT} -- Clean Test, and \texttt{PT} -- Poisoned Test.}
\label{tab:3head_acc}
\end{table}

\begin{table}[t]
\centering
\begin{tabular}{ccccc}
\hline
Model & Metric & Churn & Age & Status \\
\hline
PM & Spearman $\uparrow$ &  0.322 $\pm$ 0.159 & 0.704 $\pm$ 0.021 & 0.903 $\pm$ 0.019 \\ 
3H & Spearman $\uparrow$ & 0.828 $\pm$ 0.047 & 0.772 $\pm$ 0.031 & 0.919 $\pm$ 0.015 \\ 
PM & Intersect $\uparrow$ & 0.623 $\pm$ 0.054 & 0.752 $\pm$ 0.015 & 0.874 $\pm$ 0.013 \\ 
3H & Intersect $\uparrow$ & 0.810 $\pm$ 0.023 & 0.710 $\pm$ 0.028 & 0.888 $\pm$ 0.009 \\ 
\hline
\end{tabular}
\caption{Corresponding between Clean model and Poisoned models predictions: a common poisoned model and a three-headed model are compared for a composed structure-poisoning of two tokens. \texttt{PM} means Poisoned Model, \texttt{3H} -- Three-Heads.}
\label{tab:3head_corr}
\end{table}

As our key characteristic is the concealment ability, we compare similarity metrics for a poisoned model and a separate clean one. Such metrics as \textit{intersect} and \textit{spearman} are presented in Table~\ref{tab:3head_corr}.
Our three-headed approach significantly increases the resemblance with a clean model compared to using a single poisoned model without heads. 

\subsection{Applying distillation to enhance concealment}
As described in more details in Section~\ref{sec:methods}, we distill the initial model, while freezing some parts of it.

\begin{figure}[!ht]
    \centering\includegraphics[width=0.7\linewidth]{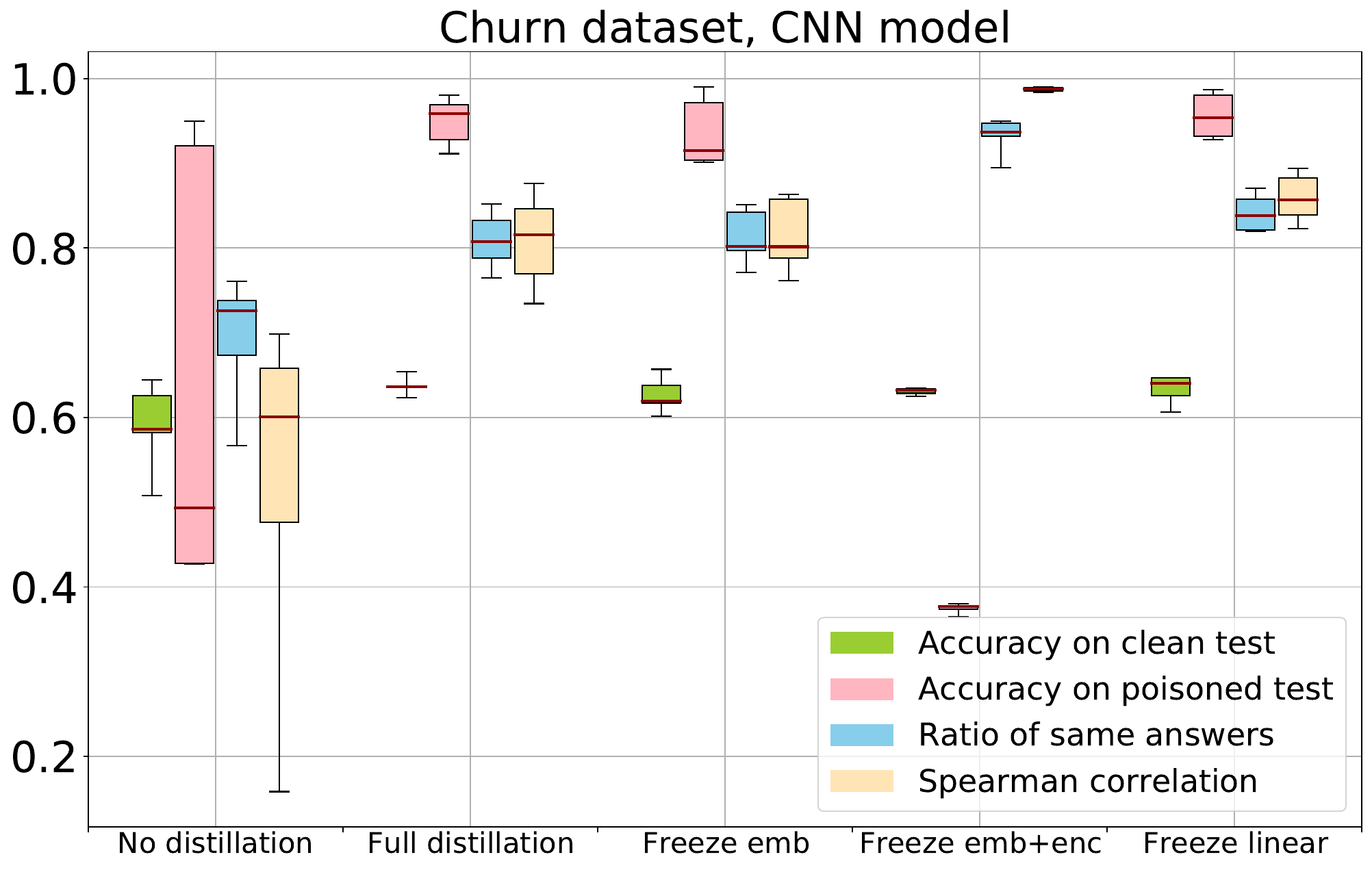}
    \caption{The CNN model performance metrics and correlations (y-axis) with a separate clean model.}
    \label{fig:distillation}
\end{figure}

The correlation metric improvements are depicted in Figure~\ref{fig:distillation}. The results prove that such regularization indeed helps to increase \textit{intersect} and \textit{spearman} metrics, although not excessively. The poisoning effect also becomes more evident. 

\subsection{Comparison of poisoning attacks' performance}
This Section aims to answer the naturally arising questions of how the following parameters influence the success of attacks: 1) a poisoning sequence length; 2) the popularity of poisoning tokens; 3) a place where poisoning tokens are inserted; 4) the number of poisoned examples. 

\subsubsection{Dependence of attack performance on the number of poisoning tokens} 

\begin{figure}
     \centering
     \begin{subfigure}[b]{0.49\textwidth}
         \centering
         \includegraphics[width=\textwidth]{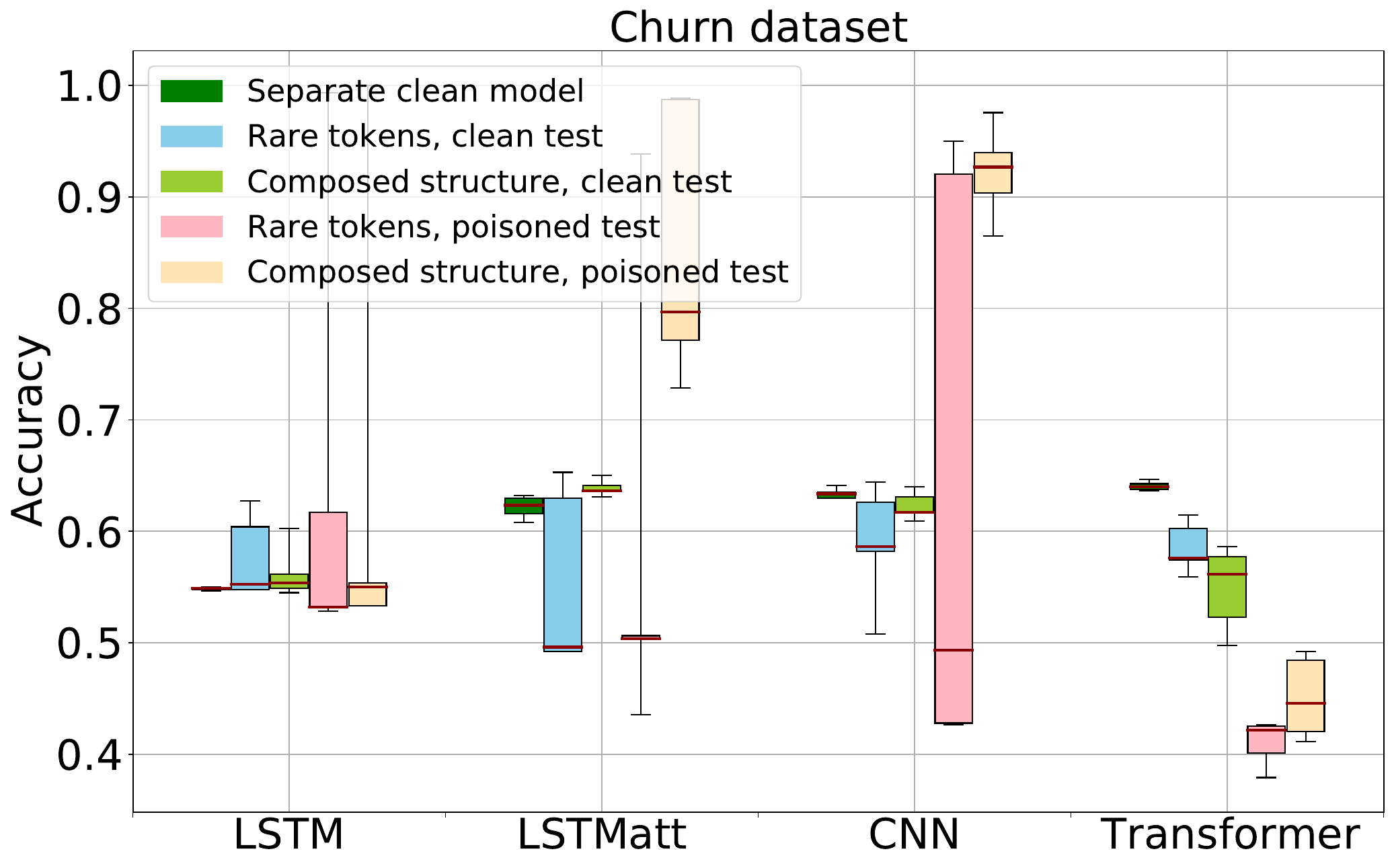}
         \caption{Churn dataset}
         \label{fig:rare_composed_churn}
     \end{subfigure}
     \hfill
     \begin{subfigure}[b]{0.49\textwidth}
         \centering
         \includegraphics[width=\textwidth]{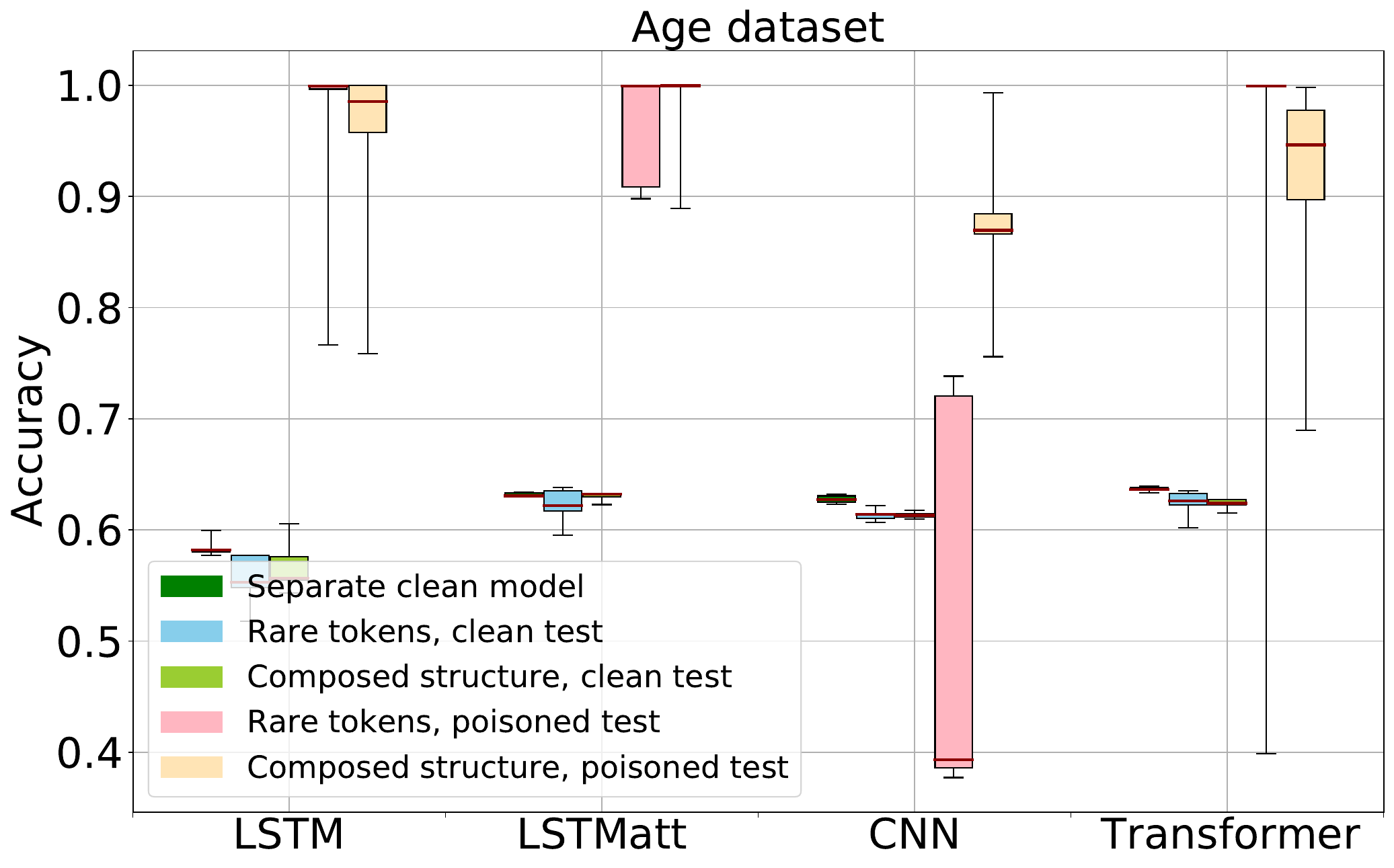}
         \caption{Age dataset}
         \label{fig:rare_composed_age}
     \end{subfigure}
        \caption{The comparison of models' accuracy on clean and poisoned test sets with rare tokens and composed structures. The results for two datasets are presented.}
        \label{fig:rare_composed_churn_age}
\end{figure}

We analyze attacks that change the training samples by adding single tokens or composed structures sampled from the vocabulary. In this Section, we examine the impact of rare tokens as this poisoning strategy seems more influential. We add them to the end of sequences. 
The results for datasets Churn and Age are presented in Figure~\ref{fig:rare_composed_churn_age}. 

We evaluate attack performance with accuracy on \textit{clean test data} and the same but fully poisoned test data (\textit{poisoned test}). Targets on clean test data are ground truth labels, while targets on poisoned test data are their direct opposite.

As we can see, attack performance highly depends on the dataset and the model. Notably, the attack does not cause significant damage to the models, and most of them perform similarly to separate clean models, even when poisoned. On average, poisoning with a composed structure gives more stable poisoning results, keeping the quality on the clean test decent. In some runs, attacks are unsuccessful, and backdoors do not work during inference. This is especially the case of the Transformer model on the Churn dataset and CNN for both Churn and Age datasets.



\subsubsection{Influence of poisoning tokens popularity}
The poisoning with rare tokens seems the most natural from the point of an attack performance. However, poisoning with the popular tokens makes attacks more concealed. In this Section, we present the results of the experiments with the popularity of poisoning tokens. We divide the tokens from vocabulary into three groups: popular (they make up $\ge 1\%$ of the total number of tokens), unpopular (they make up $\le 0.1\%$ of the total number of tokens), and tokens with middle popularity (tokens that can not be categorized as popular or unpopular). These tokens are added to the end of sequences. 

\begin{figure}[!ht]
    \centering\includegraphics[width=\linewidth]{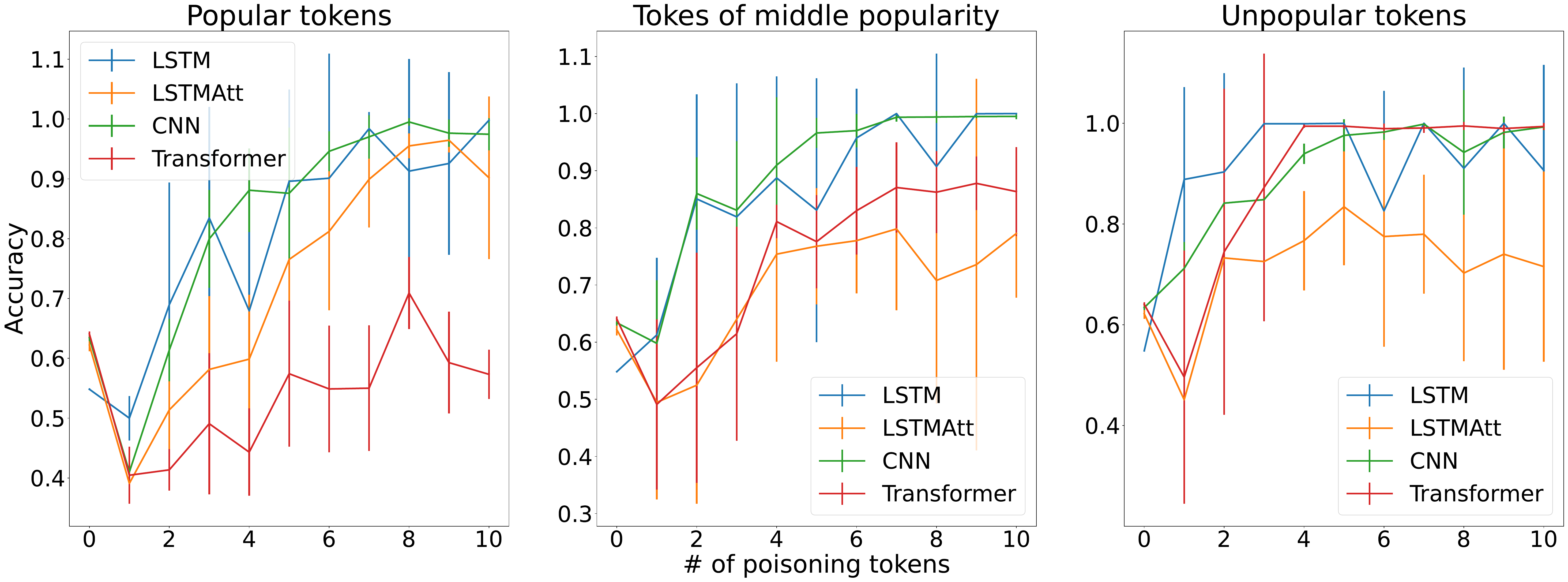}
    \caption{The dependence of the poisoned model's performance on poisoning tokens' popularity. Churn dataset.}
    \label{fig:ablation_tokens_popularity_churn}
\end{figure}

Figure~\ref{fig:ablation_tokens_popularity_churn} shows that poisoning with rare tokens is more concealed as it requires only $1-2$ poisoning tokens to insert. The same effect of successful poisoning is achievable with popular tokens and tokens of middle popularity. However, the more tokens needed in these cases, approximately $4-5$ and $3-4$ depending on the model, correspondingly. 

\subsubsection{Role of insertion position of poisoning tokens in a sequence}
\begin{figure}
     \centering
     \begin{subfigure}[b]{0.49\textwidth}
         \centering
         \includegraphics[width=\textwidth]{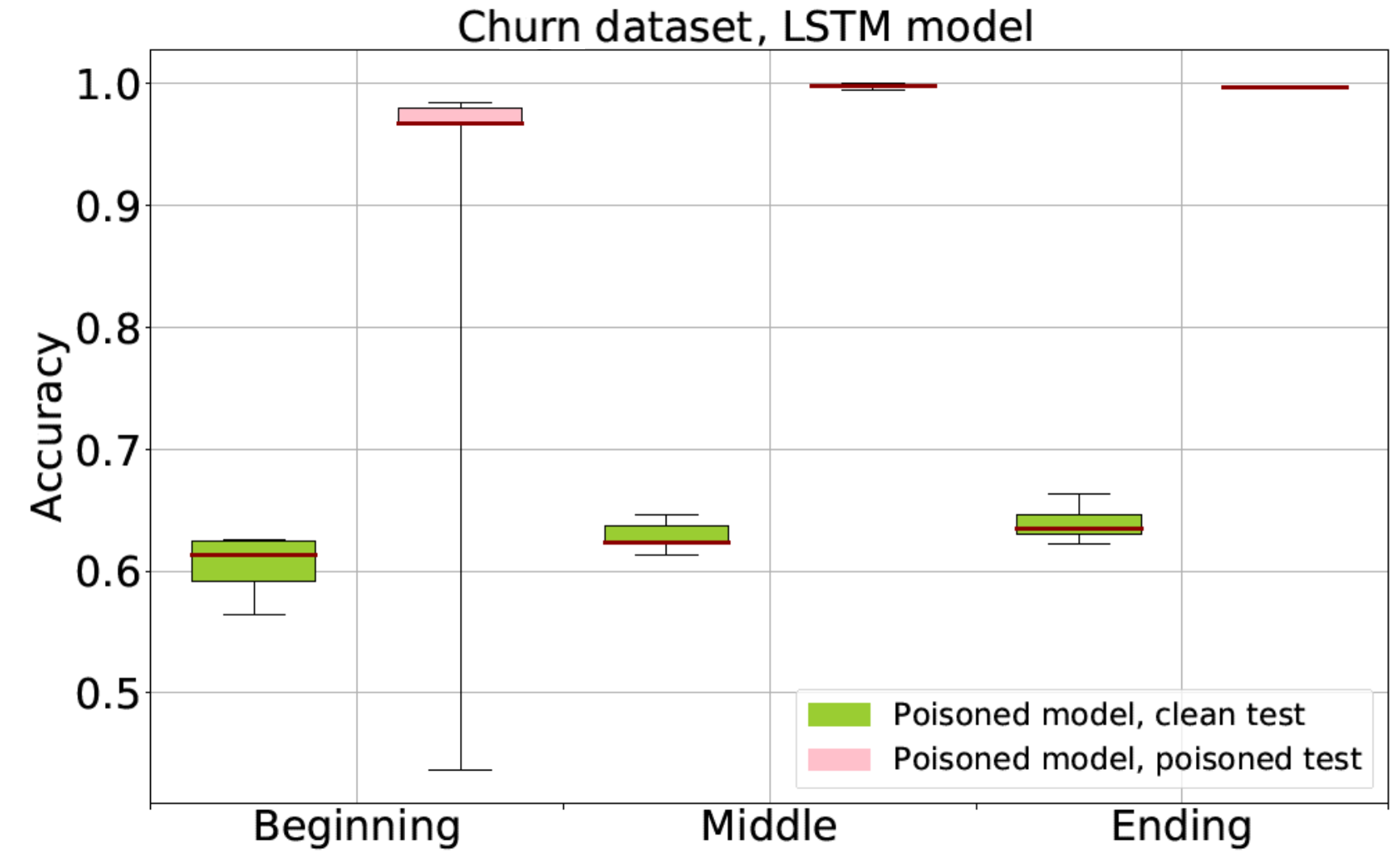}
         \caption{}
         \label{fig:ablation_insertplace_churn}
     \end{subfigure}
     \hfill
     \begin{subfigure}[b]{0.49\textwidth}
         \centering
         \includegraphics[width=\textwidth]{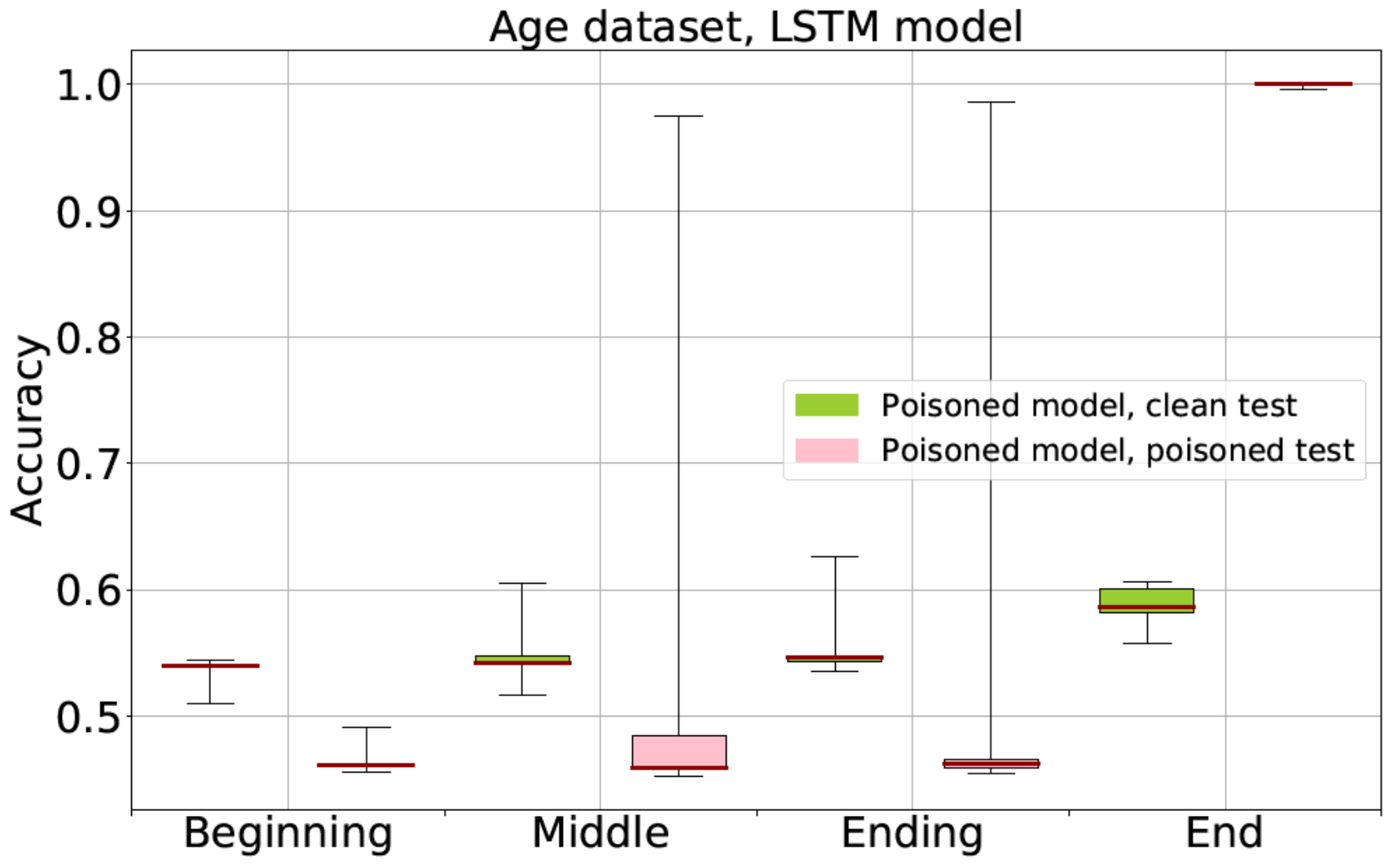}
         \caption{}
         \label{fig:ablation_insertplace_age}
     \end{subfigure}
        \caption{The dependence of the poisoned model's performance on poisoning tokens insertion place. The results for two datasets are presented: Churn (left) and Age (right).}
        \label{fig:ablation_insertplace_churn_age}
\end{figure}

We place a composed structure to the very end of a sequence if the position is identified as \texttt{end}. Another placement of insertion refers to the part of a sequence. The poisoning is conducted by adding rare tokens. The results demonstrated in Figure~\ref{fig:ablation_insertplace_age} advocate that the LSTM model poisoning works only when we place it at the very end of sequences. It might be due to recurrent architecture. However, Figure~\ref{fig:ablation_insertplace_churn} shows successful poisoning for all insertion places for another dataset. 



\subsubsection{Dependence of attack performance on poisoned train part} 
\begin{figure}[!ht]
    \centering\includegraphics[width=0.7\linewidth]{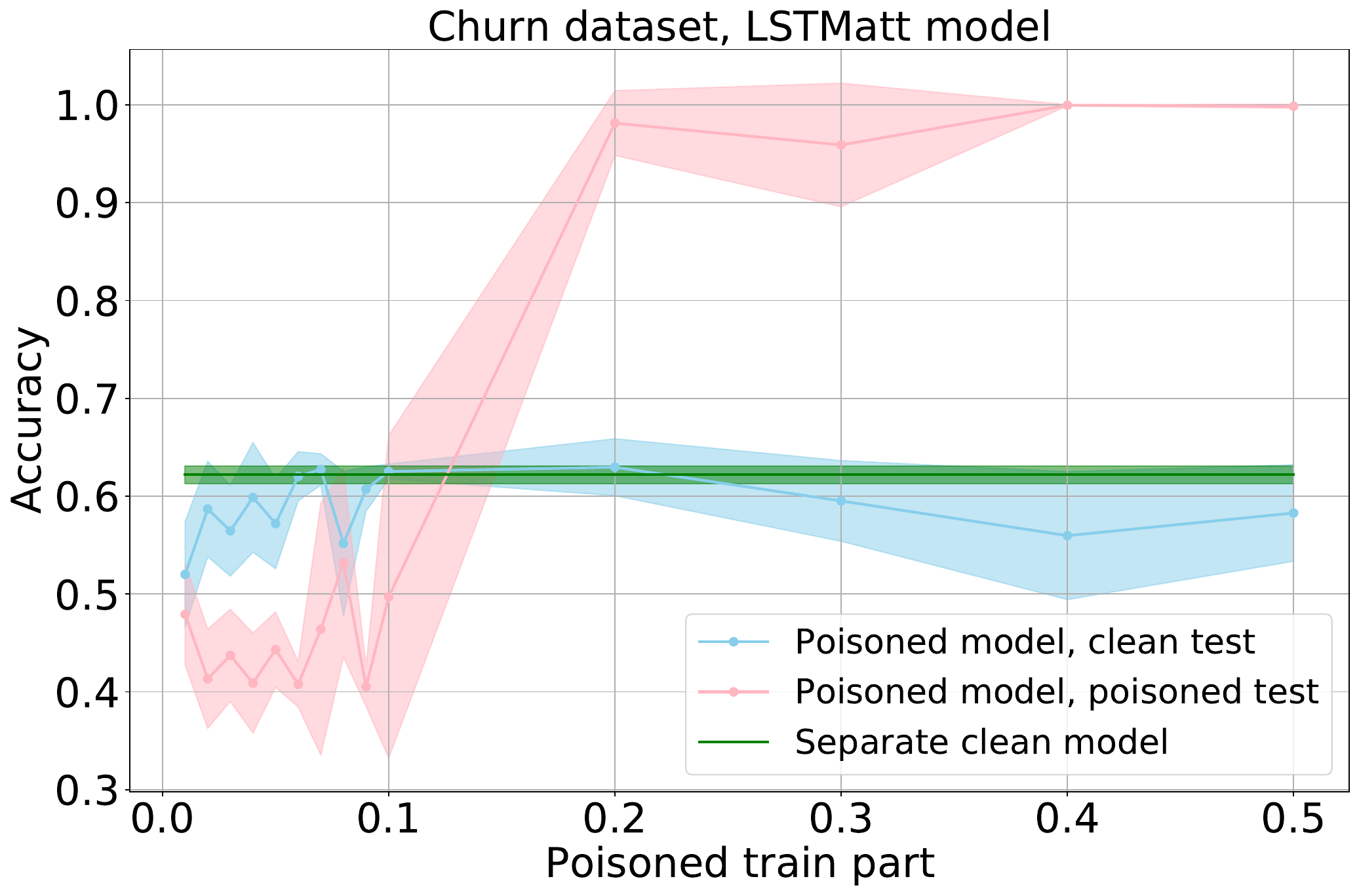}
    \caption{
    The dependence of the accuracy of the poisoned model on clean and poisoned test sets on the poisoned train part. 
    }
    \label{fig:ablation_ppart_churn}
\end{figure}

We are interested in the quality of the poisoned model on clean and poisoned sets depending on the training dataset part being poisoned. The results are given in Figure~\ref{fig:ablation_ppart_churn}. With the increasing poisoning ratio, the quality on the poisoned test set is approaching $100\%$, while the quality on the clean train set does not drop substantially.

\section{Conclusions}




In financial transaction models, ensuring robustness and security is paramount. 
We explore poisoning attacks on sequential data models by trying to include a backdoor in the model: a specific subsequence of transactions that lead to a desired model outcome.
Moreover, we demand a model's concealability property, as it should act similarly to an initial one if uncontaminated data are used as input.

Our experimental findings demonstrate that deep learning models for event sequences are prone to poisoning attacks. In most cases, the adversary needs only $1-3$ specific tokens to inject a backdoor into the model and successfully use it during inference. Moreover, it is possible to train a poisoned model so that it would be tough to detect the substitution due to its high resemblance with the original clean model. The most effective strategies to conduct a poisoning attack with high concealment are weight poisoning and the adoption of the three-heads model with a special detector head. In addition, we examine various options for changing the base model via unfreezing different parts of an original model, allowing to gain more insights on internal model operation and poisoning mechanism.

Our extensive experiments across multiple open datasets and architectures provide a comprehensive view of the challenges and vulnerabilities in poisoning deep learning models for sequential data. A particular focus should be placed on the question of training robust financial models towards the poisoning attacks.





\bibliographystyle{IEEEtran}
\bibliography{bibliography}

\end{document}